\documentclass{ecai} 

\usepackage{latexsym}
\usepackage{amssymb}
\usepackage{amsmath}
\usepackage{amsthm}
\usepackage{booktabs}
\usepackage{enumitem}
\usepackage{graphicx}
\usepackage{color}
\usepackage{subfigure}
\usepackage{multirow}
\usepackage{bm}

\newtheorem{theorem}{Theorem}

\newcommand{\BibTeX}{B\kern-.05em{\sc i\kern-.025em b}\kern-.08em\TeX}

\begin{document}


\begin{frontmatter}


\paperid{372} 


\title{Fully Hyperbolic Rotation for Knowledge Graph Embedding}


\author[A,B,C]{\fnms{Qiuyu}~\snm{Liang}}
\author[A,B,C]{\fnms{Weihua}~\snm{Wang}\thanks{Corresponding Author. Email: wangwh@imu.edu.cn.}}
\author[A,B,C]{\fnms{Feilong}~\snm{Bao}}
\author[A,B,C]{\fnms{Guanglai}~\snm{Gao}}

\address[A]{College of Computer Science, Inner Mongolia University, Hohhot, China}
\address[B]{National and Local Joint Engineering Research Center of Intelligent Information Processing \\ Technology for Mongolian, Hohhot, China}
\address[C]{Inner Mongolia Key Laboratory of Multilingual Artificial Intelligence Technology, Hohhot, China}


\begin{abstract}
Hyperbolic rotation is commonly used to effectively model knowledge graphs and their inherent hierarchies.
However, existing hyperbolic rotation models rely on logarithmic and exponential mappings for feature transformation.
These models only project data features into hyperbolic space for rotation, limiting their ability to fully exploit the hyperbolic space.
To address this problem, we propose a novel fully hyperbolic model designed for knowledge graph embedding.
Instead of feature mappings, we define the model directly in hyperbolic space with the Lorentz model.
Our model considers each relation in knowledge graphs as a Lorentz rotation from the head entity to the tail entity.
We adopt the Lorentzian version distance as the scoring function for measuring the plausibility of triplets.
Extensive results on standard knowledge graph completion benchmarks demonstrated that our model achieves competitive results with fewer parameters.
In addition, our model get the state-of-the-art performance on datasets of CoDEx-s and CoDEx-m, which are more diverse and challenging than before. 
Our code is available at https://github.com/llqy123/FHRE.

\end{abstract}

\end{frontmatter}


\section{Introduction}

Knowledge Graphs (KGs) store a vast amount of facts in the form of triplets, which provide a rich external semantic knowledge base for many artificial intelligence tasks, such as word sense disambiguation \citep{ALMOUSA2022101337}, question answering \citep{chen-etal-2023-multi} and recommendation systems \citep{10.1145/3583780.3615110}.
Due to the diverse range of applications, representation learning for knowledge graphs has attracted significant interest from researchers.

The construction of knowledge graphs typically involves automatically extracting triplets from multiple sources of knowledge to form a graphical representation.
They usually comprise numerous entities, relations and an extensive collection of triplets.
However, since these KGs are automatically extracted from the knowledge sources, they are incomplete \citep{le2023knowledge}.
In the context of large-scale knowledge graphs, it is impractical to complete missing facts manually. 
To address this challenge, researchers have proposed several algorithms for automatic filling, Knowledge Graph Completion (KGC), also known as link prediction. 
Its aim is to the prediction of missing facts using known triples.
In KGC, a common approach to predict missing facts in KGs is to embed them into vector spaces, which is called Knowledge Graph Embedding (KGE).
The KGE model calculates the valid facts with higher scores through their scoring function.

\begin{figure}[!htb]
\centering
\includegraphics[width=0.9\linewidth]{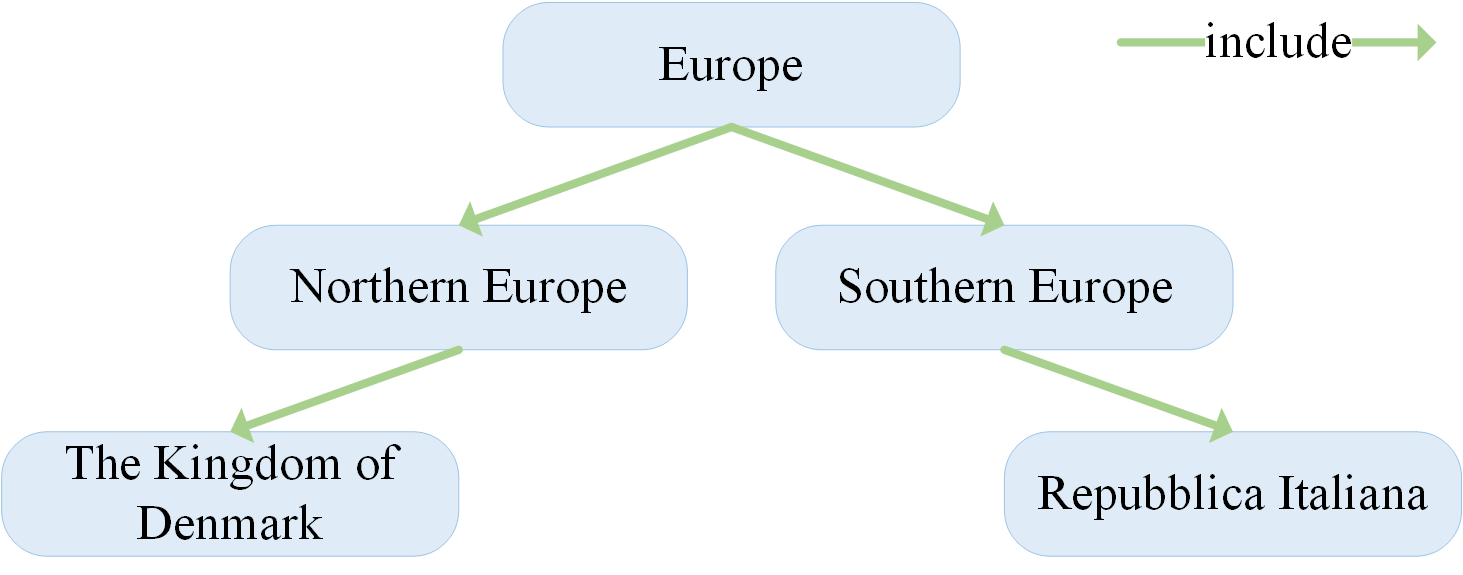}
\caption{The example of the European entity illustrates how KG presents hierarchies.}
\label{fig:hie}
\end{figure}

In terms of the complexity of neural networks, KGE models can be categorized into shallow network-based models \citep{cao2022geometry,gao2020rotate3d,le2023knowledge,ZEB2024120197,zhang2019quaternion} and deep network-based models \citep{10.1007/978-3-031-33455-9_16,Zhang2023MultiAspectEC}.
Shallow network-based models based on the principle of geometric transformation are becoming increasingly popular among researchers, as these models are lightweight and interpretable. 
They model the KG mainly through geometric operations such as translation, rotation and reflection. 
Based on these advantages, our paper focuses on shallow networks.

It is well known that real-world knowledge graphs tend to exhibit hierarchical structures, as illustrated in Figure \ref{fig:hie}.
In Figure~\ref{fig:hie}, the concept of ``Europe'' encompasses both ``Northern Europe'' and ``Southern Europe''.
The Northern European countries could include ``the Kingdom of Denmark'' and the Southern European countries could include ``Repubblica Italiana''.
However, Euclidean space inherently assumes a flat and uniform structure.
It is challenging for Euclidean space to represent and distinguish between different levels when embedding hierarchical structures.

Fortunately, hyperbolic space \citep{balazevic2019multi, ganea2018hyperbolic, liang20242} can be regarded as continuous analogue of discrete trees, which endows them with the natural ability to capture hierarchical structures.
For example, MuRP \citep{balazevic2019multi}, RotH \citep{chami2020low} and FFTRotH \citep{xiao2022complex-fft} borrow the hierarchical modeling capabilities of hyperbolic spaces.
However, a common disadvantage of current hyperbolic models is that they frequently rely on logarithmic and exponential mappings to transform data features between hyperbolic and tangent spaces. 
In other words, each epoch and batch of data must undergo spatial transformation through mapping functions during the training phase.
These hybrid models only project data features into hyperbolic space to perform entity and relation transformations, limiting their ability to fully utilize hyperbolic space. 
Specifically, logarithmic and exponential mappings consist of a series of hyperbolic and inverse hyperbolic functions.
These functions are highly complex and often have infinite range, which weakens the stability of the model.

To address this, we propose a \textbf{F}ully \textbf{H}yperbolic \textbf{R}otation model for knowledge graph \textbf{E}mbedding, named as \textbf{FHRE}.
Specifically, we define the model directly in hyperbolic space with the Lorentz model instead of spatial mappings.  
Then, our model considers each relation in knowledge graph as a Lorentz rotation from the head entity to the tail entity.
Finally, we adopt the Lorentzian version distance as a scoring function for measuring the plausibility of triplets.

Our contributions then become four-fold:
\begin{itemize}
    \item To the best of our knowledge, we are the first to propose modelling knowledge graphs using Lorentz rotations in a fully hyperbolic space.
    \item We propose a novel hyperbolic rotation model, named \textbf{FHRE}, effectively solves the problem that hyperbolic rotation models rely on spatial mappings.
    \item Experimental results on standard knowledge graph benchmarks (FB15k-237 and WN18RR) demonstrate that our model achieves competitive results in both low and high dimensions.
    \item We further validate the effectiveness and generalization ability of our approach on more diverse and challenging benchmark datasets (CoDEx-s and CoDEx-m). 
    Experiments show that our model get the state-of-the-art performance compared to the latest model.
    
\end{itemize}


\section{Related Work}

The embedding space of shallow network-based models can be classified into three main categories: Euclidean, Complex and Hyperbolic.

\textbf{Euclidean-based embeddings.}
TransE \citep{bordes2013translating} treated the relation vectors of the knowledge graph as translations from head entities to tail entities.
Although TransE is fairly simple and has few parameters, it cannot model knowledge graphs with one-to-many, many-to-one and many-to-many relationships.
To overcome these shortcomings, a series of translation models (e.g., TransH \citep{wang2014knowledge} and TransR \citep{lin2015learning}) were proposed.
While these models perform excellently, they are unable to model the rich logical patterns (e.g., symmetry, inversion and composition) in knowledge graph.
To solve these problems, rotation models based on Euclidean space have been proposed, such as RotE \citep{chami2020low} and recent CompoundE \citep{ge2023compounding} model.
However, using Euclidean space with a constant curvature of 0, which is a flat space, is insufficient to accurately model the hierarchical structure of a knowledge graph.
Thus resulting in distorted data.

\textbf{Complex-based embeddings.}
In order to improve the representation of space, researchers have tried to model knowledge graphs in complex spaces.
Entities and relations in complex space consist of real and imaginary parts.
For example, RotatE \citep{sun2019rotate} defined each relation in complex space as a rotation from the head entity to the tail entity.
QuatE \citep{zhang2019quaternion} explored embeddings in hypercomplex spaces and used Hamilton products for rotation operations.
Rotate4D \citep{le2023knowledge} decomposed the relation into unit vectors to perform 4D rotations in a hypercomplex space.
Although these models achieve better results, one drawback of these embeddings is that they often require high-dimensional spaces and then increased memory costs.

\textbf{Hyperbolic-based embeddings.}
In recent years, due to the development of hyperbolic geometry in the field of artificial intelligence, more and more researchers use the properties of hyperbolic geometry to model the knowledge graph.
For example, MuRP \citep{balazevic2019multi} was the first to focus on the hierarchical structure of KGs and transformed entity embeddings by learning relation-specific parameters via Mobius matrix vector multiplication and Mobius addition.
RotH \citep{chami2020low} performed rotation modelling knowledge mapping in hyperbolic space.
HBE \citep{pan2021hyperbolic} used the extended Poincaré ball and polar coordinate system to capture hierarchies.
FFTRotH \citep{xiao2022complex-fft} further enhanced the RotH model by a Fast Fourier Transform.
The latest model CoPE \citep{ZEB2024120197} learned embeddings using the Poincaré ball of hyperbolic geometry to preserve the hierarchy between entities.
Although these models address the issue of hierarchical structure, they only project data features into hyperbolic space, limiting their ability to fully leverage the potential of hyperbolic space.
Thus resulting in a lack of performance in knowledge graph embedding.

\begin{table}
\renewcommand\arraystretch{1}
\caption{Mathematical symbols used in this paper.}
\label{tab:sym}
\centering
\begin{tabular}{cc} 
\toprule
Symbols & Descriptions\\
\hline
$\mathcal{G}$ & Knowledge Graph\\
$\mathcal{V}$ & the set of entities  \\
$\mathcal{R}$ & the set of relations  \\
$\mathbf{h}$, $\mathbf{r}$, $\mathbf{t}$ &  head entity, relation, tail entity  $\mathbf{h}, \mathbf{t} \in \mathcal{V}$, $\mathbf{r} \in \mathcal{R}$   \\
$\mathbb{R}$ & Real number  \\
$\mathbb{H}$ & Hyperbolic space  \\
$c$ & space curvature  \\
 $\mathcal{L}^n_c$&  $n$-dimensional Riemannian manifold \\
$\mathcal{T}_{\textbf{x}}\mathcal{L}^n_c$ & the tangent space at point $\mathbf{x}$ \\
$\left \langle ,\right \rangle_\mathcal{L}$ &  Lorentzian scalar product  \\
$\mathbf{v_h}$,$\mathbf{v_t}$ & head and tail entity embeddings in hyperbolic space
  \\
$\mathbf{\theta_r}$ & parameterized relation embeddings  \\
$\mathbf{{v_h}^{\prime}}$ &  head embedding that have undergone Lorentz rotation  \\
\bottomrule
\end{tabular}
\end{table}

\begin{figure*}[!htb]
\centering 
\subfigure[Rotation in Euclidean space]{
\label{Fig:eurota}
\includegraphics[width=0.32\linewidth]{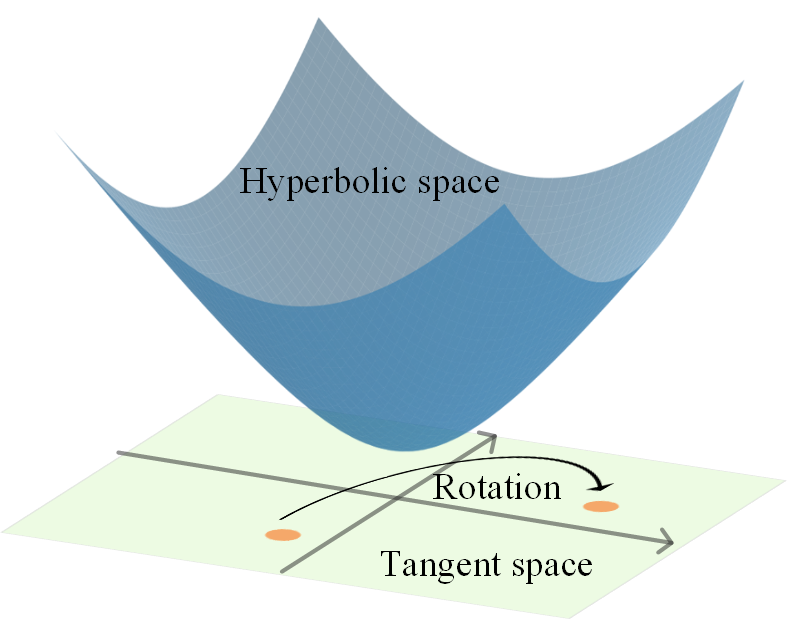}}\subfigure[Rotation in hyperbolic space with the logarithmic and exponential mappings]{
\label{Fig:hyperrota}
\includegraphics[width=0.32\linewidth]{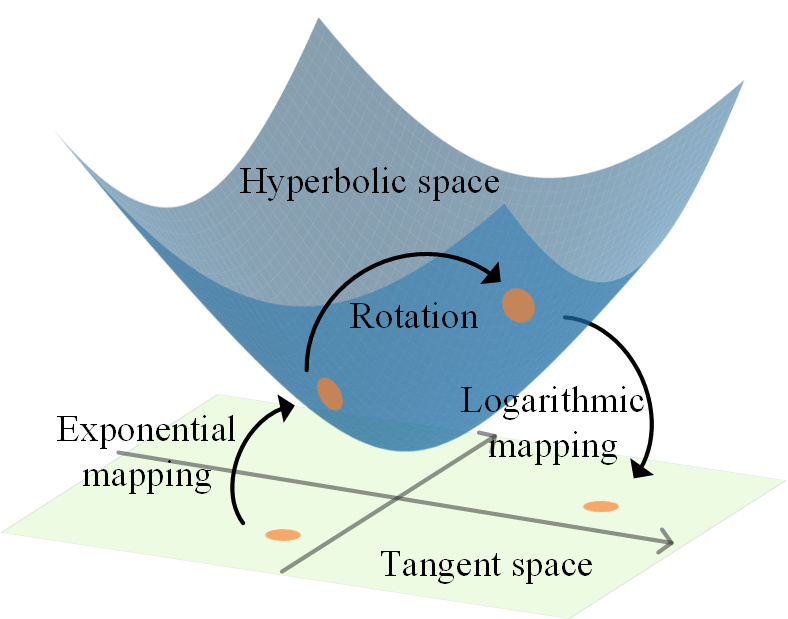}}\subfigure[Rotation in fully hyperbolic space]{
\label{Fig:fullyeurota}
\includegraphics[width=0.32\linewidth]{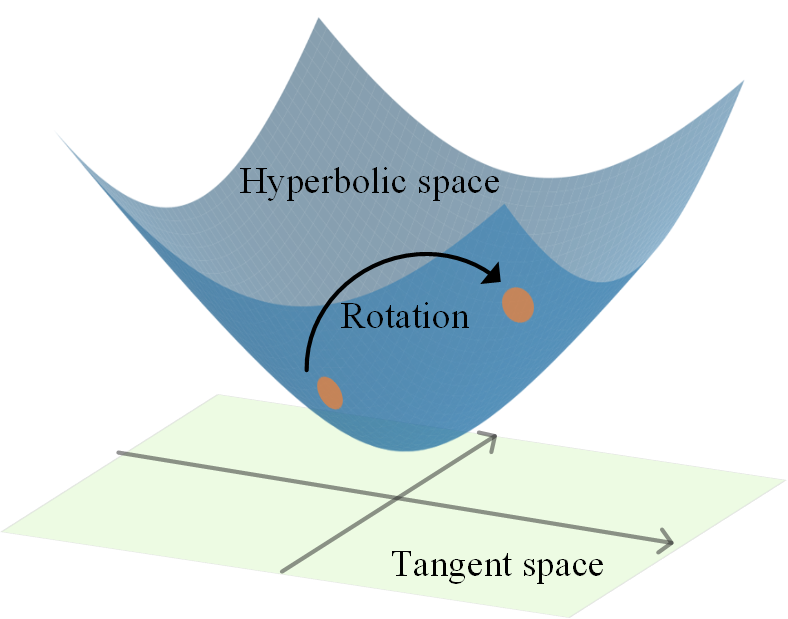}}
\caption{Rotational transformations based on different spaces in the training phase.
Orange dot indicates entity in knowledge graph.
Tangent space as subspace of Euclidean space.
}
\label{Fig:compare}
\end{figure*}

\section{Problem Formulation and Preliminaries}

In this section, we first introduce the definition of knowledge graph completion task, and then provides a brief background of the background knowledge of hyperbolic geometry.
Table \ref{tab:sym} provides a summary of the main mathematical symbols utilized throughout the paper.

\subsection{Knowledge graph completion}
For a knowledge graph $\mathcal{G}$, we represent each piece of data in $\mathcal{G}$ in the form of a triplet $(\mathbf{h},\mathbf{r},\mathbf{t})$.
In the task of knowledge graph completion, there are three main categories: head entity completion $(?,\mathbf{r},\mathbf{t})$, tail entity completion $(\mathbf{h},\mathbf{r},?)$ and relation completion $(\mathbf{h},?,\mathbf{t})$.
Our paper will focus on head or tail completion, as is the case in most previous work. 
This is because models require relational information to be trained.

\subsection{Hyperbolic geometry}
Hyperbolic geometry is a non-Euclidean geometry with constant negative curvature $c$.
A lower value of $c$ typically indicates a more curved surface.
Previous research has demonstrated the effectiveness of hyperbolic geometries, such as the Poincaré ball \citep{ganea2018hyperbolic} and the Lorentz model \citep{nickel2018learning}, in the field of natural language processing. 
Our model is based on the Lorentz model due to its simplicity and numerical stability \citep{pmlr-v202-mishne23a}.

\textbf{Lorenz model.}
The Lorentz model of $n$-dimensional hyperbolic space is defined as the Riemannian manifold $\mathcal{L}^n_c = (\mathcal{H}^n, g_\ell )$, where $g_\ell = \text{diag}(-1,1,\dots,1)$ is the Riemannian metric tensor and manifold $\mathcal{H}^n$ satisfies:
\begin{equation}
\mathbf{x} \in \mathbb{R}^{n+1}: \left \langle \mathbf{x},\mathbf{x}\right \rangle_\mathcal{L}=-1,
\end{equation}
where $\left \langle ,\right \rangle_\mathcal{L}$ is Lorentzian scalar product.
Given $\mathbf{x}, \mathbf{y} \in \mathbb{R}^{n+1}$, where $n$ is the space dimension.
Formally, the Lorentzian scalar product is defined as
\begin{equation}
\begin{aligned}
    \left \langle \mathbf{x},\mathbf{y}\right \rangle_\mathcal{L} &=-x_0y_0+\displaystyle\sum_{i=1}^{n}{x_ny_n} \\
    & = \mathbf{x}^T\text{diag}(-1,1,\dots,1)\mathbf{y}.
\end{aligned}
\end{equation}

In addition, when describing points in the Lorentz model, we also follow the terminology of special relativity \citep{Bdeir2024, chen2022fully}.
That means taking the first dimension to be the time component $x_0$ and its remaining dimensions to be the space component $x_n$.
Thus, each point in $\mathcal{L}^n_c$, has the form $\mathbf{x}=[x_0,\mathbf{x}_n]^T$ and $x_0=\sqrt{||\mathbf{x_n}||^2-\frac{1}{c}}$, where $c$ is the curvature of hyperbolic space.
In the latter sections, we denote a point $\mathbf{x}$ in the Lorentz model as $\mathbf{x} \in \mathcal{L}^n_c$.

\textbf{Tangent Space.}
Given $\mathbf{x} \in \mathcal{L}^n_c$, the tangent space at that point is defined:
\begin{equation}
    \mathcal{T}_{\mathbf{x}}\mathcal{L}^n_c =\{\mathbf{y} \in \mathbb{R}^{n+1}|\left \langle \mathbf{y},\mathbf{x}\right \rangle_{\mathcal{L}}=0\}.
\end{equation}
Note that a tangent space is a type of Euclidean space.

\textbf{Logarithmic and Exponential Mappings.}
The logarithmic mapping is used to map features from hyperbolic space to tangent space, while the exponential mapping is used to map features from tangent space to hyperbolic space.
Therefore, given tangent vector $\mathbf{z} \in \mathcal{T}_{\mathbf{x}}\mathcal{L}^n_c$ to $\mathcal{L}^n_c$ by moving along the geodesic $\gamma $, where $\gamma (0)=\mathbf{x}$ and $\gamma^{\prime} (0)=\mathbf{z}$.
The exponential mapping $\text{exp}^c_{\mathbf{x}}(\mathbf{z})$ process can be defined as follows:
\begin{equation}
\begin{aligned}
\label{equ:exp}
   \mathcal{T}_{\mathbf{x}}\mathcal{L}^n_c \Rightarrow \mathcal{L}^n_c : \text{exp}^c_{\mathbf{x}}(\mathbf{z}) &=\text{cosh}(\alpha )\mathbf{x}+\text{sinh}(\alpha )\frac{\mathbf{z}}{\alpha },\\
   \alpha &=\sqrt{-c}||\mathbf{z}||_\mathcal{L},\\
   ||\mathbf{z}||_\mathcal{L} &=\sqrt{\left \langle \mathbf{z},\mathbf{z}\right \rangle}_\mathcal{L}.
\end{aligned}
\end{equation}
Conversely, given hyperbolic vector $\mathbf{y} \in \mathcal{L}^n_c$, the logarithmic $\text{log}^c_{\mathbf{x}}(\mathbf{y})$ mapping to $\mathcal{T}_{\mathbf{x}}\mathcal{L}^n_c$ is defined as follows:
\begin{equation}
\begin{aligned}
    \mathcal{L}^n_c \Rightarrow \mathcal{T}_{\mathbf{x}}\mathcal{L}^n_c : \text{log}^c_{\mathbf{x}}(\mathbf{y}) &=\frac{\text{cos}^{-1}(\beta )}{\sqrt{\beta^2 -1}}(\mathbf{y}-\beta \mathbf{x}),\\
    \beta &=c\left \langle \mathbf{x},\mathbf{y}\right \rangle_\mathcal{L},
\end{aligned}
\end{equation}
where $\gamma (0)=\mathbf{x}$ and $c$ is the curvature of the hyperbolic space.
In the next work, for simplicity, we set $c$ to 1, i.e., the curvature is -1.

\textbf{Lorentzian distance}.
Given $\mathbf{x},\mathbf{y} \in \mathcal{L}^n_c$, the Lorentzian distance can be defined as the following equation:
\begin{equation}
\label{equ:dis}
    d^2_\mathcal{L}(\mathbf{x},\mathbf{y})=\frac{2}{c}-2\left \langle \mathbf{x},\mathbf{y}\right \rangle_\mathcal{L},
\end{equation}
where $d^2_\mathcal{L}(\mathbf{x},\mathbf{y}) \in \mathcal{L}^n_c$, and $c$ is the curvature of hyperbolic space.

\section{Methodology}

In this section, we will introduce the detail of our model.
Our model is comprises of three key components:
\begin{itemize}
    \item \textbf{Initialization}: Our model initializes entity and relation embeddings in hyperbolic space.
    \item \textbf{Lorentz rotation}: Our model treats each relation as a rotation that from the head entity to the corresponding candidate tail entity. 
    The rotation guided the head entities to stay closer with tail entities. 
    The rotation operations are performed fully in a hyperbolic space. 
    \item \textbf{Scoring function}: The prediction step calculates the scoring function to measure the plausibility of a triplet. 
    We propose the Lorentzian distance as our scoring function.
\end{itemize}

\subsection{Initialization}

To obtain the embedding in the hyperbolic space, we randomly initialize the head and tail entity embedding parameters in tangent space, and initialize parameterized relation embedding as $\mathbf{\theta_r}$.
Then we map these entity embeddings to hyperbolic space using exponential mapping (Equation \ref{equ:exp}) and get their embedding of the head entities $\mathbf{v_h}$ and tail entities $\mathbf{v_t}$ in the hyperbolic space.
It is noting that our model only requires once spatial transformation during the initialization process and does not depend on mapping with other parts of the process.

\subsection{Lorentz rotation}
\label{Lorentz rotation}

The use of rotational transformation has been proven to efficiently encode complex logical structures in knowledge graphs, including symmetric, anti-symmetric, inverse and compositional relations \cite{chami2020low,gao2020rotate3d,sun2019rotate, xiao2022complex-fft}.

As an example, in Figure \ref{Fig:compare}, we show rotational transformations based on different spaces.
Figure \ref{Fig:eurota} shows the rotation in Euclidean space.
The rotational transformation of an entity does not depend on hyperbolic space.
Figure \ref{Fig:hyperrota} illustrates the rotation in hyperbolic space. 
To perform a rotational transformation in hyperbolic space, an entity in tangent space (Euclidean space) must be mapped into hyperbolic space by exponential mapping. 
This is followed by a mapping of the transformed entities back into tangent space by logarithmic mapping for training purposes.
Please note that the same conversion process must be applied to each batch of data.

Figure \ref{Fig:fullyeurota} illustrates the rotation in full hyperbolic space, which is the method we propose.
The initialized data is projected into hyperbolic space by an exponential mapping function, without relying on any mapping function for data feature transformation during the training phase.

In our approach, we propose that each relation can be considered as a Lorentz rotation from the head to the tail entity embedding.
The Lorentz rotation theorem is as follows:
\begin{theorem}[Lorentz rotation]
Lorentz rotation is the rotation of the spatial coordinates. 
The Lorentz rotation matrix is of the following form:
\begin{equation}
    \mathrm{R}=\left [ \begin{matrix}1
 & 0\\ 
 0& \mathrm{\widetilde{R}}
\end{matrix}\right ],
\end{equation}
where $\mathrm{\widetilde{R}^T}\mathrm{\widetilde{R}}=\mathrm{I}$ and $\mathbf{det}(\mathrm{\widetilde{R}})=1$.
$\mathrm{\widetilde{R}} \in \mathrm{SO}(n)$ is a special orthogonal matrix.
\end{theorem}

The Lorentz rotation is the linear transformations directly defined in the Lorentz model, i.e., $\forall \mathbf{x} \in \mathcal{L}^n_c,  \mathrm{R}\mathbf{x} \in \mathcal{L}^n_c$.

Denote $\theta $ as the angle of rotation. 
The rotation matrix $\mathrm{\widetilde{R}}$ can represent the rotation transformation, which is defined as follows:
\begin{equation}
    \mathrm{\widetilde{R}}(\theta )=
    \begin{bmatrix}
    cos(\theta ) & - sin(\theta )\\ 
    sin(\theta ) &  cos(\theta )
    \end{bmatrix},
\end{equation}
where $\mathrm{\widetilde{R}^T}(\theta )\mathrm{\widetilde{R}}(\theta )=\textbf{I}$ and $\textbf{det}\;\mathrm{\widetilde{R}}(\theta )=1$.

Let $\mathbf{\theta_r}=(\theta_{r,j})_{j\in\{1,\cdots ,\frac{d}{2}\}}$ denote as the parameterized relation embeddings, where $d$ denotes an even number of embedding dimension.
The rotation operation can be defined as:
\begin{equation}
    Rot(\theta_r)=\text{diag}(\mathrm{\widetilde{R}} (\theta _{r,1}),\cdots ,\mathrm{\widetilde{R}} (\theta _{r,\frac{d}{2}})),
\end{equation}
where $ Rot(\theta_r) \in \mathcal{L}^n_c$.

Thus, given a head entity embedding $\mathbf{v_h}$ and relation embedding $\mathbf{\theta_r}$, the process of Lorentz rotation is defined as follows:
\begin{equation}
\label{equ:giv}
   \mathbf{{v_h}^{\prime}} = \mathbf{v_h} \otimes \mathbf{\theta_r} = \mathbf{v_h}Rot(\theta_r),
\end{equation}
where $ \mathbf{{v_h}^{\prime}} \in \mathcal{L}^n_c$,
``$\otimes$'' represents the Lorentz rotation.

\subsection{Scoring function}

Similar to previous studies \citep{balazevic2019multi,chami2020low,xiao2022complex-fft}, we adopt the distance function as our scoring function to measure the plausibility of triplets.
The score for each triplet $(\mathbf{h},\mathbf{r},\mathbf{t})$ is defined as:
\begin{equation}
    s(\mathbf{h},\mathbf{r},\mathbf{t})= d^2_\mathcal{L}(\bm{\mathbf{{v_h}^{\prime}}},\bm{\mathbf{v_t}}) + b_h + b_t,
\end{equation}
where $d^2_\mathcal{L}(,)$ is the Lorentzian distance (Equation \ref{equ:dis}), $\bm{\mathbf{v_h}}^{\prime}$ is the head entity after Lorentz rotation (Equation \ref{equ:giv}), $\bm{\mathbf{v_t}}$ is the hyperbolic embedding of the tail entity, $b_h$ and $b_t$ are entity biases which act as margins in the scoring function \citep{chami2020low, tifrea2018poincare}.

\subsection{Loss function}
Following with previous work \citep{balazevic2019multi,chami2020low, chen2022fully}, for each triplet, we randomly corrupt its head or tail entity with $k$ entities and compute the probability of the triplet as $p=\sigma(s(\mathbf{h},\mathbf{r},\mathbf{t}))$, where $\sigma$ is the sigmoid function.
Finally, we train our model by minimizing the binary cross entropy loss:
\begin{equation}
    loss = -\frac{1}{N}\displaystyle\sum_{i=1}^{N}(\text{log}(p^{(i)})+\displaystyle\sum_{j=1}^{k}\text{log}(1-\widetilde{p}^{(i,j)})),
\end{equation}
where $N$ is the number of training set triplets, $p^{(i)}$ and $\widetilde{p}^{(i,j)}$ are the probabilities of correct and incorrect triplets, respectively.

\begin{table}
\renewcommand\arraystretch{1}
\setlength{\tabcolsep}{5.5pt}
\centering
\caption{Data statistic on five datasets.}
\label{tab:datastat}
\begin{tabular}{ccccccc}
\hline
\textbf{Dataset} & Entities &  Relations  &  Train  & Valid   & Test                               \\ 
\hline
FB15k-237        & 14,541   & 227  &272,115      & 17,535     &20,466   \\
WN18RR           & 40,943   & 11  &86,835       & 3,034      &3,134     \\
CoDEx-s         & 2,034  & 42   & 32,888         & 1,827  &1,828 \\
CoDEx-m         & 17,050  & 51   &  185,584      & 10,310  &10,310 \\
Nations         & 14  & 55   & 1,592      & 199 &201 \\
\hline
\end{tabular}
\end{table}

\begin{table*}
\renewcommand\arraystretch{1.0}
\caption{
Link prediction results on FB15k-237 and WN18RR for \textbf{low-dimensional} embeddings ($d$ = 32) in the filtered setting.
$\mathbb{E}$, $\mathbb{C}$ and $\mathbb{H}$ are denoted as the presentation space of Euclidean, complex and hyperbolic space, respectively.
Best results are in bold and second best results are underlined (The same settings are applied to Table \ref{high} and Table \ref{newres}).
$\clubsuit$ indicates that the results are from RotH \citep{chami2020low}.
}

\setlength{\tabcolsep}{6pt}
\centering
\normalsize
\label{lower}
\begin{tabular}{clccccccccc}
\hline
\multirow{2}{*}{\textbf{Space}} &\multirow{2}{*}{\textbf{Model}} & \multicolumn{4}{c}{FB15k-237}  &   & \multicolumn{4}{c}{WN18RR}    \\ 
\cline{3-6}
\cline{8-11}
 &    & MRR     & H@1  & H@3 & \multicolumn{1}{c}{H@10} &    & MRR     & H@1  & H@3 & \multicolumn{1}{c}{H@10}         \\ 
\hline
\multirow{4}{*}{$\mathbb{E}$} 
&MuRE$^\clubsuit$   & 0.313  &0.226   &0.340  &0.489  &  & 0.458  &0.421   &0.471  &0.525     \\ 
&RotE & 0.307  &0.220   &0.337  &0.482 &    & 0.463  &0.426   &0.477  &0.529           \\ 
&Rot2L & 0.326  &0.237   &-  &0.503  &  & 0.475  &0.434   &-  &0.554               \\ 
&SAttLE-Tucker & \underline{0.340}  &\underline{0.252} &\underline{0.372} &0.513  & & 0.454  &0.414   &0.474  &0.527            \\ 

\cline{2-11}
\multirow{2}{*}{$\mathbb{C}$} 
&RotatE$^\clubsuit$   & 0.290  &0.208   &0.316  &0.458  &   & 0.378  &0.330   &0.417  &0.491         \\ 
&ComplEx-N3$^\clubsuit$  & 0.294  &0.211   &0.322  &0.463  & & 0.420  &0.390   &0.420  &0.460      \\
\cline{2-11}

\multirow{6}{*}{$\mathbb{H}$} 
&MuRP$^\clubsuit$    & 0.323  &0.235  &0.353  &0.501  & & 0.465  &0.420   &0.484  &0.544           \\ 
&RotH   & 0.314 &0.223   &0.346  &0.497  & & 0.472  &0.428   &0.490  &0.553          \\ 
&FFTRotH & 0.319  &0.228   &0.352  &0.500  &  & 0.484  &0.437   &0.502  &\textbf{0.572}          \\ 
&HYBONET & 0.334  &0.244   &0.365  &\underline{0.516}  &  & \underline{0.489}  &\underline{0.455}   &\underline{0.503} &0.553  \\ 
&UltraE & 0.334  &0.243   &0.360  &0.510  &  & 0.488  &0.440   &0.503 &0.558  \\ 
\cline{2-11}
&\textbf{FHRE}(ours) &\textbf{0.345}	&\textbf{0.255}	&\textbf{0.375}	&\textbf{0.528}  & &\textbf{0.494} &\textbf{0.458}	&\textbf{0.510}	&\underline{0.563}		 \\

\hline
\end{tabular}
\end{table*}

\section{Experiments}
\subsection{Experimental settings}

\textbf{Datasets}.
To evaluate the effectiveness of our model, we used two frequently benchmark datasets, FB15k-237 \cite{toutanova2015observed} and WN18RR \cite{dettmers2018convolutional}.
The FB15k-237 and WN18RR datasets are subsets of the FB15k \cite{bordes2013translating} and WN18 \cite{dettmers2018convolutional} datasets, respectively. 
They were created to address the issue of reversible relations, and enabling more realistic predictions. 

Furthermore, we also validated the robustness of our model on CoDEx-s and CoDEx-m \cite{safavi2020codex}.
The CoDEx-s and CoDE-m datasets were proposed in \citep{safavi2020codex} to enlarge the KG scope and improve the level of KG difficulty.
This dataset includes three knowledge graphs with different sizes and structures.
Importantly, it contains thousands of hard negative triples that are plausible but verified to be false.
These two datasets are more diverse and interpretable benchmarks.
Therefore, the CoDEx dataset is a more difficult link prediction benchmark than FB15k-237.

Finally, we also evaluated our model on Nations \cite{10.5555/1597538.1597600} dataset.
In this dataset, the number of entities is smaller than the number of relation.
Table \ref{tab:datastat} provides statistics for all datasets.
For a fair comparison, we used the same partition of train, valid and test with other work.

\textbf{Evaluation Protocols}.
Similar to previous work \citep{chami2020low, chen2022fully,lacroix2018canonical}, we augment all datasets by adding inverse relations to each triplet.
In other words, we add an additional triplet $(\mathbf{t},\mathbf{r}^{-1},\mathbf{h})$ for every $(\mathbf{h},\mathbf{r},\mathbf{t})$.
We adopt the Mean Reciprocal Rank (MRR) and Hits@k (k=1, 3 or 10) as evaluation metrics.
Higher MRR and Hits@k values on the valid set indicate better model performance. 
The final scores on the test set are obtained from the best validation model, which achieved the highest MRR on the validation set.

\textbf{Baselines}.
In the low-dimensional experiments, we compared with 10 baseline models.
These models can be classified into three categories depending on the embedding space:
\begin{itemize}
    \item \textbf{Euclidean space:} 
    MuRE \citep{balazevic2019multi} transformed entity embeddings by learning relation-specific parameters in Euclidean space. 
    RotE \citep{chami2020low} performed rotation operations in Euclidean space. 
    Rot2L \citep{wang2021hyperbolic} modelled knowledge graphs in Euclidean space using a double-layer superposition transformation.
    SAttLE-Tucker \citep{BAGHERSHAHI2023110124} utilized a large number of self-attention heads as the key to applying query-dependent projections to capture mutual information between entities and relations.
    \item \textbf{Complex space}: 
    RotatE \citep{sun2019rotate} and ComplEx-N3 \cite{lacroix2018canonical} modelled knowledge graphs in complex space to improve their expression ability.
    \item \textbf{Hyperbolic space}: 
    MuRP \citep{balazevic2019multi} transformed entity embeddings by learning relation-specific parameters in Poincaré ball.
    RotH \citep{chami2020low} performed rotation operations in hyperbolic space. 
    FFTRotH \citep{xiao2022complex-fft} improved RotH model with Fast Fourier Transform. 
    HYBONET \citep{chen2022fully} performed a Lorentz linear transformation on each triplet in hyperbolic space.
    UltraE \citep{10.1145/3534678.3539333} presented an ultra-hyperbolic KG embedding method that interleaves hyperbolic and spherical manifolds. 
    Note that MuRP, RotH, FFTRotH and UltraE models rely on spatial mappings between hyperbolic space and their tangent space.

\end{itemize}

In the high-dimensional experiments, we compared more baselines besides to the above models.
They are:
HAKE \citep{zhang2020learning} modelled the hierarchical structure of knowledge graphs using polar coordinate systems.
CompoundE \citep{ge2023compounding} modelled KGs using translation, rotation and scaling operations to form a new combined operation.
QuatE \citep{zhang2019quaternion} unified ComplEx-N3 and RotatE by modeling relations as rotations on quaternion space.
Rotate4D \citep{le2023knowledge} treated relations as 4D rotations from head to tail entities in quaternion space.
GIE \citep{cao2022geometry} proposed a knowledge graph embedding model based on geometric space interaction.
HBE \citep{pan2021hyperbolic} used the extended Poincaré ball and polar coordinate system to capture hierarchies.

In the CoDEx-s, CoDEx-m and Nation datasets, we compared CompGCN \citep{vashishth2020compositionbased}, WGE \citep{10.1007/978-3-031-33455-9_16}, NoGE \citep{10.1145/3488560.3502183}, ATTH \citep{chami2020low} and the latest model CoPE \citep{ZEB2024120197}.
The CompGCN, WGE and NoGE models are based on deep neural networks. \footnote{Since models based on shallow networks are less reported on these datasets.}
ATTH model combined rotations and reflections for KGs in hyperbolic space.
CoPE model learned embeddings using the Poincaré ball of hyperbolic geometry to preserve the hierarchy between entities.

\begin{table}
\caption{The optimal hyper-parameters on the five datasets with different embedding dimensions.
$d$ and $e$ indicate the embedding dimension and training epoch.
}
\label{tab:para}
\renewcommand\arraystretch{1}
\setlength{\tabcolsep}{4.5pt}
\centering
\begin{tabular}{ccccccc}
\hline
\textbf{Para.}    & \multicolumn{2}{c}{FB15k-237}  & \multicolumn{2}{c}{WN18RR}    & CoDEx-\{s,m\}  & Nations  \\ 
\hline
$d$  &32 &500  &32 &500 &64  &32\\
$lr$   & 5e-3     & 5e-3  &5e-3 &3e-3    &5e-3  & 5e-3     \\ 
$neg$  & 50       & 50   &100  &200   &10  & 10  \\
$b$       & 500    & 500  &1000 &1000    &128  & 128     \\    
$e$       & 800    & 800   &1000 &1000     &500  & 500  \\ 
\hline
\end{tabular}
\end{table}

\begin{table*}[ht]
\renewcommand\arraystretch{1.0}
\caption{
Link prediction results on FB15k-237 and WN18RR for \textbf{high-dimensional} embeddings ($d \in \{200,300,500,1000\}$) in the filtered setting.
$\mathbb{E}$, $\mathbb{C}$, $\mathbb{Q}$, $\mathbb{M}$ and $\mathbb{H}$ are denoted as the representation space of Euclidean, complex, quaternion, mixed and hyperbolic, respectively.
$\clubsuit$ indicates that the results are from RotH \citep{chami2020low}.
}
\setlength{\tabcolsep}{6pt}
\centering
\label{high}
\normalsize
\begin{tabular}{clccccccccc}
\hline
\multirow{2}{*}{\textbf{Space}} &\multirow{2}{*}{\textbf{Model}}  & \multicolumn{4}{c}{FB15k-237} & & \multicolumn{4}{c}{WN18RR}        \\ 
\cline{3-6}
\cline{8-11}
 &    & MRR     & H@1  & H@3 & \multicolumn{1}{c}{H@10} &    & MRR     & H@1  & H@3 & \multicolumn{1}{c}{H@10}         \\ 
\hline
\multirow{4}{*}{$\mathbb{E}$} 
&MuRE$^\clubsuit$    & 0.336  &0.245   &0.370  &0.521 & & 0.475  &0.436   &0.487  &0.554         \\ 
&RotE                & 0.346  &0.251   &0.381  &0.538  & & 0.494  &0.446   &0.512  &0.585           \\ 
&HAKE                & 0.346  &0.250   &0.381  &0.542 &  & 0.497  &0.452   &0.516  &0.582           \\ 
&CompoundE  & 0.357  &0.264   &0.393  &0.545 &  & 0.491  &0.450   &0.508  &0.576       \\ 

\cline{2-11}
\multirow{2}{*}{$\mathbb{C}$} 
&RotatE$^\clubsuit$   & 0.338  &0.241   &0.375  &0.533  &  & 0.476  &0.428   &0.492  &0.571      \\ 
&ComplEx-N3$^\clubsuit$ & 0.357  &0.264 &0.392  &0.547  &  & 0.480  &0.435   &0.495  &0.572      \\
\cline{2-11}

\cline{2-11}
\multirow{2}{*}{$\mathbb{Q}$} 
&QuatE   & 0.348  &0.248   &0.382  &0.550   & & 0.488  &0.438   &0.508  &0.582          \\ 
&Rotate4D& 0.353  &0.257   &0.391  &0.547   &  & 0.499  &0.455   &0.518  &\underline{0.587}     \\
\cline{2-11}

\multirow{1}{*}{$\mathbb{M}$} 
&GIE   & \underline{0.362} &\underline{0.271}   &\underline{0.401}  &0.552  &  & 0.491  &0.452   &0.505  &0.575          \\ 
\cline{2-11}

\multirow{6}{*}{$\mathbb{H}$} 
&MuRP$^\clubsuit$    & 0.335  &0.243  &0.367  &0.518  & & 0.481  &0.440   &0.495  &0.566          \\ 
&RotH   & 0.344 &0.246   &0.380  &0.535 &   & 0.496  &0.449   &0.514  &0.586           \\ 
&HBE   & 0.336 &0.239   &0.372  &0.534 &  & 0.488  &0.448   &0.502  &0.570          \\ 
&HYBONET& 0.352 &0.263   &0.387  &0.529  &  & \underline{0.513}  &\textbf{0.482}   &\underline{0.527} &0.569         \\ 
&UltraE & 0.351  &0.275   &0.400  &\textbf{0.560}  &  & 0.501  &0.450   &0.515 &0\textbf{.592}  \\ 
\cline{2-11}
&\textbf{FHRE}(ours)&\textbf{0.374}	&\textbf{0.281}	&\textbf{0.409}	&\underline{0.558} &  &\textbf{0.515}	&\underline{0.478}	&\textbf{0.529}	&0.586		 \\

\hline
\end{tabular}
\end{table*}

\begin{table*}
\renewcommand\arraystretch{1.0}
\caption{
Link prediction results on the CoDEx-s, CoDEx-m and Nations dataset.
$\mathbb{E}$ and $\mathbb{H}$ are Euclidean and hyperbolic space, respectively.
$\Diamond$  and $\heartsuit $ indicate that the results are from NoGE \citep{10.1145/3488560.3502183} and CoPE \citep{ZEB2024120197}, respectively.
}
\setlength{\tabcolsep}{5pt}
\label{newres}
\centering
\normalsize
\begin{tabular}{clccccccccccc}
\hline
\multirow{2}{*}{\textbf{Space}} &\multirow{2}{*}{\textbf{Model}}  & \multicolumn{3}{c}{CoDEx-s}  & & \multicolumn{3}{c}{CoDEx-m} & & \multicolumn{3}{c}{Nations}     \\ 
\cline{3-5}
\cline{7-9}
\cline{11-13}
 &    & MRR     & H@1  & \multicolumn{1}{c}{H@10} &  & MRR     & H@1  & \multicolumn{1}{c}{H@10} &   & MRR     & H@1   & \multicolumn{1}{c}{H@10}         \\ 
\hline
\multirow{3}{*}{$\mathbb{E}$} 
&CompGCN$^\Diamond$   & 0.395  &-  &0.621  & & 0.312  &- &0.457  & & -  & -  & -    \\ 
&WGE$^\Diamond$    & 0.452  &- &\underline{0.664}  & & \underline{0.338}  &- &\underline{0.485}  & & -  & -  & -    \\ 
&NoGE$^\Diamond$        & \underline{0.453}  &-  &0.650  & & \underline{0.338}  &- &0.484  & & -  &-  &-    \\ 
\cline{2-13}

\multirow{4}{*}{$\mathbb{H}$} 
&MuRP$^\heartsuit $    & 0.420  &0.311   &0.632    & & 0.306  &0.226   &0.456  & & 0.818  &0.726   &\textbf{1.0}    \\ 
&ATTH$^\heartsuit $    & 0.402  &0.286   &0.632    & & 0.315 &0.237  &0.464 & & 0.785  &0.664   &0.995       \\ 
&CoPE$^\heartsuit $    & 0.446  &\underline{0.350} &0.631  & & 0.326 &\underline{0.251} &0.466  & & \underline{0.835}  &\underline{0.754}   &\textbf{1.0}     \\ 
\cline{2-13}
&\textbf{FHRE}(ours) &\textbf{0.598}	&\textbf{0.513}	&\textbf{0.765}	&	&\textbf{0.391}	&\textbf{0.316}		&\textbf{0.536} &	&\textbf{0.885}	&\textbf{0.838}		&\underline{0.997} \\

\hline
\end{tabular}
\end{table*}

\textbf{Implementation Details}.
In our experiments, we conducted a grid search to select optimal hyper-parameters, for learning rate $lr$ $\in$ \{1e-3, 3e-3, 5e-3\}, negative sample size $neg$  $\in$ \{10, 50, 100, 200\} and batch size $b$ $\in$ \{128, 500, 1000\}.
We optimized our model with Riemannian Adam \citep{kochurov2020geoopt} and conducted all our experiments on an NVIDIA Tesla P100 GPU with 16GB memory.
In low-dimensional settings, the embedding dimension of all experiments was set to 32, while the embedding dimension was set to 500 in high-dimensional settings.
The optimal hyper-parameters of our model on these datasets are detailed in Table \ref{tab:para}.

\subsection{Results}

The spatial accommodation capacity of hyperbolic spaces grows exponentially, which means that models based on hyperbolic spaces could perform better even in lower dimensions. 
Therefore, we first compared our model with other models when the embedding dimension is 32.
Then, we compare our models with those models with higher dimensions.

For the FB15k-237 and WN18RR datasets, we show the experimental results for low-dimensional embeddings in Table \ref{lower} and for high-dimensional embeddings in Table \ref{high}.
For the CoDEx-s, CoDEx-m and Nation datasets, we show the experimental results in Table \ref{newres}.
The results of other model we compared are from the original papers.

\subsubsection{Low-dimensional embedding experiments}
Table \ref{lower} demonstrates that our model outperforms all other competitors on the FB15k-237 and WN18RR datasets.
In the same dimensions ($d=32$), our FHRE outperforms the Euclidean space-based model Rot2L by an average of 6.1\% and 3.7\% on the FB15k-237 and WN18RR datasets, respectively.
We argue that although Rot2L also uses rotations to model knowledge graphs, our FHRE is superior to it.
Our rotation was operated in hyperbolic space, which allows FHRE to capture hierarchical structures more efficiently.

In comparison to models based on hyperbolic spaces, both RotH and FFTRotH utilize hyperbolic rotations to model the knowledge graph.
However, these models project data features into hyperbolic space for rotational transformation during training, which limits their ability to fully exploit the hyperbolic space and results in poor performance.
Our model is based on a fully hyperbolic space and does not rely on frequent spatial transformations and thus shows excellent performance on both datasets.
We even outperform the UltraE model since it learned the embeddings with an ultrahyperbolic manifold.

Although HYBONET \citep{chen2022fully} performs a Lorentzian linear transformation in hyperbolic space, the linear transformation introduces additional parameters.
Our analysis is as follows: given the same embedding dimension $d$, HYBONET has $(|\mathcal{V}| \times d + |\mathcal{R}| \times d \times d)$ parameters, while our model has $(|\mathcal{V}| \times d +|\mathcal{R}| \times d)$ parameters, where $\mathcal{V}$ and $\mathcal{R}$ are the set of entities and relations, respectively. 
Therefore, according to the linear transformation time complexity calculation method, the time complexity of our model is $\mathbf{O(n)}$, while the time complexity of HYBONET is $\mathbf{O(n^2)}$.
Moreover, our experimental observations have confirmed that rotational transformations are more expressive than linear transformations for modelling knowledge graphs. 
We also observed that increasing the number of parameters does not necessarily result in a corresponding performance improvement.

\subsubsection{High-dimensional embedding experiments}
Table \ref{high} illustrates that our model achieve superior performance compared to other models with high embedding dimension, particularly on the FB15k-237 dataset.
Although the HAKE utilized polar coordinates to model the hierarchical structure of the knowledge graph, it is modelled in Euclidean space and cannot fully capture the hierarchical structure.
Our model even exceed in the mixed space model GIE.
GIE combined Euclidean, hyperbolic and spherical spaces to form a new interaction space when modeling knowledge graphs.
However, GIE encountered spatial transformation issues.

While our metrics for the WN18RR model are not optimal in terms of H@1 and H@10, our model has fewer parameters compared to Rotate4D and HYBONET.
Table \ref{tab:paranum} illustrates the number of parameters.
For a fair comparison, we fixed the embedding dimension at 500.
Table \ref{tab:paranum} clearly demonstrates that our model's parameters are 21.1\% and 74.9\% lower than those of the Rotate4D and HYBONET models, respectively.

\begin{table}[ht]
\caption{500-dimensional embedding parameter comparison.} 
\label{tab:paranum}
\renewcommand\arraystretch{1.1}
\centering
\begin{tabular}{lc}
\hline
\textbf{Model}  &  Number of parameters           \\ 
\hline
    Rotate4D    & 81.9 \textbf{M}     \\
    HYBONET     & 26.0 \textbf{M}      \\
    FHRE(ours)  & 20.5 \textbf{M}       \\
\hline 
\end{tabular}
\end{table}

\subsubsection{Robustness and generalization experiments}
To validate the robustness and generalization of our model, we performed experiments on more diverse and challenging datasets of different scales.
The table \ref{newres} shows that our model achieves better results compared to other models.
Compared to CoPE model, our model improved the MRR, H@1 and H@10 metrics by 34.0\%, 46.5\%, 21.2\%, 19.9\%, 25.8\%, 15.0\%, 5.9\% and 11.1\% on the CoDEx-s, CoDEx-s and Nations datasets.
Although CoPE employed the Poincaré ball model of hyperbolic space to preserve the hierarchies between entities, the CoPE encountered frequent spatial mapping problems.

Our model outperforms deep network models such as NoGE, achieving better results. 
We believe that NoGE, based on deep neural networks, is susceptible to over-fitting issues. 
Furthermore, as shown in Table \ref{newres}, the experiments demonstrate that our model maintains excellent performance in the face of more diverse and challenging datasets, which verifies the robustness of our model and its ability to perform well in a variety of situations.

\subsection{Exploring the multi-relations}

To assess the effectiveness of our model when handling with multi-relation(1-to-N, N-to-1 and N-to-N) between entities.
According to the calculation rules in \citep{bordes2013translating},  we divided the test set of FB15k-237 into four categories: 1-to-1, 1-to-N, N-to-1 and N-to-N.
``1-to-N'' means that a head entity can form a fact triplet with multiple tail entities.
``N-to-1'' means that a tail entity can form a fact triplet with multiple head entities.
The division results are shown in Table \ref{tab:categories}, where $\eta_h$ and $\eta_t$ represent the average degree of head and tail entities, respectively.

\begin{table}[t]
\renewcommand\arraystretch{1}
\caption{Classification rules and results for different relation types on FB15k-237. }
\label{tab:categories}
\centering
\begin{tabular}{cccc}
\hline
 \multicolumn{1}{c}{\textbf{Categories}}        & $\eta_h$   & $\eta_t$     & Number of triplets        \\ 
\hline
\multicolumn{1}{c}{1-1}          & $<1.5$     & $<1.5$       & 725     \\ 
\multicolumn{1}{c}{1-N}         & $<1.5$     & $>1.5$       & 19    \\ 
\multicolumn{1}{c}{N-1}         & $>1.5$     & $<1.5$       & 19577    \\ 
\multicolumn{1}{c}{N-N}        & $>1.5$     & $>1.5$       & 145    \\ 
\hline
\end{tabular}
\end{table}

We compared our model with HYBONET\citep{chen2022fully} model when fixed the embedding dimension for entities and relations at 32.
For a fair comparison, we reproduced the HYBONET with the optimal hyper-parameter settings from their paper, and remained other setting same with ours.
The experiment results are shown in Table \ref{tab:multi-relation}.

For Table~\ref{tab:multi-relation}, we observe that our model is effective when deal with different multi-relation entities.
Compared with the HYBONET model's linear transformation, we attribute this performance gain to the Lorentz rotation in hyperbolic space are more expressive when modeling complicated relations between entities.

\begin{table}
\caption{Result of each relation category on FB15k-237 for HYBONET and our model.
Best results are in bold.
}
\label{tab:multi-relation}
\renewcommand\arraystretch{1.05}
\setlength{\tabcolsep}{2.5pt}
\centering
\begin{tabular}{cccccccccc}
\hline
\multirow{2}{*}{\textbf{Categories}} &\multicolumn{4}{c}{HYBONET} & & \multicolumn{4}{c}{FHRE(ours)}        \\ 
\cline{2-5}
\cline{7-10}
 & MRR     & H@1  & H@3 & \multicolumn{1}{c}{H@10}  &  & MRR     & H@1  & H@3 & \multicolumn{1}{c}{H@10}         \\ 
\hline
1-1     &\textbf{0.240}     &\textbf{0.185} &\textbf{0.260} &0.345  & &0.239 &0.177  &0.258  &\textbf{0.346} \\ 
1-N     &0.340              &0.315          &0.315          &0.394  & &\textbf{0.369} &\textbf{0.342} &\textbf{0.368} &\textbf{0.421}\\ 
N-1    &0.335 &0.244  &0.366  &0.516  & &\textbf{0.348} &\textbf{0.255}  &\textbf{0.382}  &\textbf{0.534}  \\ 
N-N  &0.371 &\textbf{0.248} &0.448 &0.593 & &\textbf{0.373} & 0.244 &\textbf{0.455} &\textbf{0.606}\\ 
\hline
\end{tabular}
\end{table}

\begin{figure}[!htb]
\centering 
\subfigure[RotH]{
\label{Fig:sub:RotH}
\includegraphics[width=0.45\linewidth]{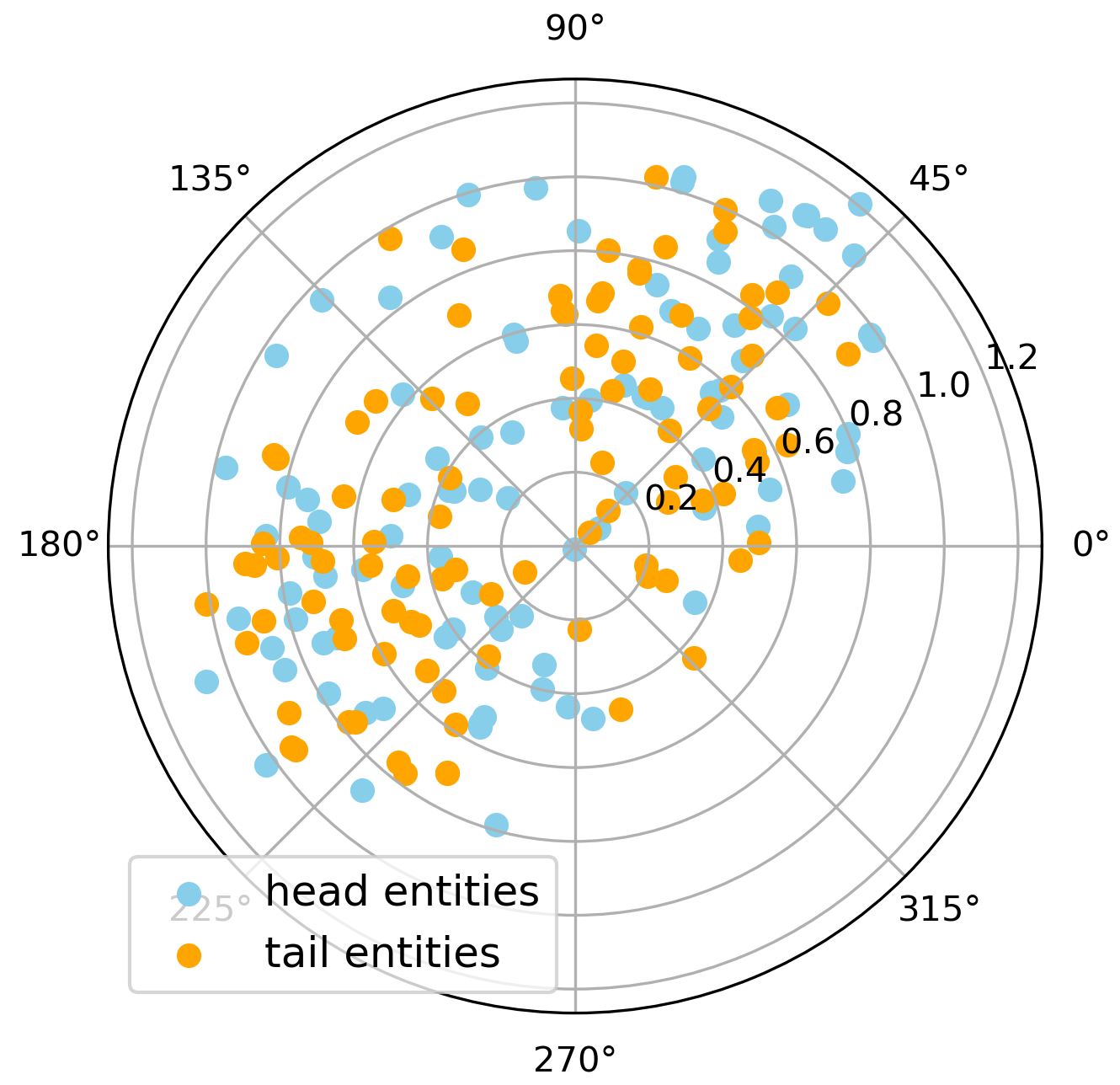}}
\subfigure[FHRE (ours)]{
\label{Fig:sub:FHRE}
\includegraphics[width=0.45\linewidth]{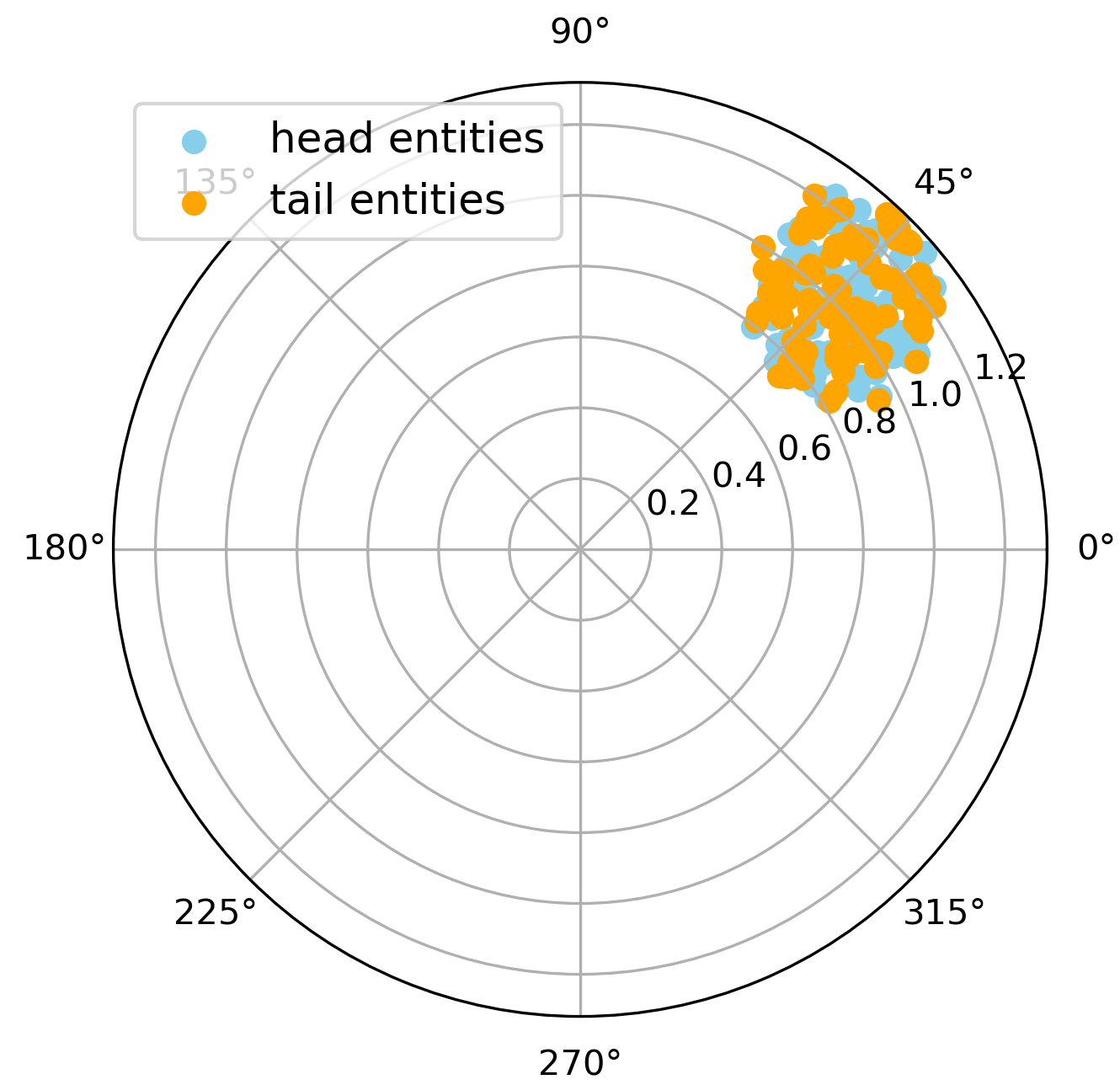}}
\caption{Embedding of the entities learned with rotation on relation \emph{\textbf{\_derivationally\_related\_form}} by RotH model and our model . 
}
\label{Fig:compare2}
\end{figure}

\subsection{Visualizations of Rotation in hyperbolic space}
To verify the impact of rotations entirely in hyperbolic space, we visualized the same embedding after rotation under performed our model and RotH model on the WN18RR dataset.
The dimensions of embeddings are 32.
We selected the first 1000 triplet instances associated with the relation ``\emph{\textbf{\_derivationally\_related\_form}}'' and used the t-SNE method to reduce the dimension of head and tail entity embeddings.
We then projected them onto a polar coordinate system for visualization. 
With this relation, entities expects to be closer and together.

Figure \ref{Fig:compare2} shows the embedded visualisations.
In Figure \ref{Fig:sub:RotH}, RotH performs rotation in hyperbolic space.
In Figure \ref{Fig:sub:FHRE}, our model performs rotation in fully hyperbolic space.
It is evident that the entity in our model is closely associated with its associated entities.
However, in the RotH, the entity is relatively isolated from other entities.
This discrepancy can be attributed to the frequent spatial transformations of the data features when training.
The spatial transformations between different representation space could lead to lose the semantic and structure information of original KGs during the transformation process.
Our model is rotation entirely in a hyperbolic space without spatial transformation, which compensates for this limitation.

\section{Conclusion}

In this paper, we propose FHRE, a fully hyperbolic rotation model for knowledge graph embedding.
In contrast to previous hyperbolic KGE models, our model does not employ exponential and logarithmic mappings to transform data features during the training phase. 
Instead, it is defined directly on the Lorenz model.
We conducted extensive experiments on five datasets.
In the standard benchmarks WN18RR and FB15k-237, our model has demonstrated competitive results in both low and high dimensions.
We conducted an analysis of the time complexity of the model, which indicated that our model could be faster.
Furthermore, we have validated the effectiveness and generalization ability of our approach on more diverse and challenging benchmark datasets.
The experiment demonstrates that our model get the state-of-the-art performance compared to the latest model.
Finally, visualization results show that FHRE can encourage entities with similar semantics to have similar embeddings, which is beneficial to the prediction of unknown triplets.


\begin{ack}
This work is supported by National Natural Science Foundation of China (Nos.62066033, 61966025);
Inner Mongolia Natural Science Foundation (Nos.2024MS06013, 2022JQ05); 
Inner Mongolia Autonomous Region Science and Technology Programme Project (Nos.2023YFSW0001, 2022YFDZ
0059,2021GG0158);
Collaborative Innovation Project between Universities and Institutes in Hohhot;
Inner Mongolia University Technology Research Fund (No. 21221505);
the fund of Supporting the Reform and Development of Local Universities (Disciplinary Construction) and the special research project of First-class Discipline of Inner Mongolia of China (YLXKZX-ND-036);
We also thank all anonymous reviewers for their insightful comments.
\end{ack}


\bibliography{mybibfile}

\end{document}